# Parameter Efficient Fine Tuning Llama 3.1 for Answering Arabic Legal Questions: A Case Study on Jordanian Laws

Mohammed Fasha
Business Intelligence and Data Analytics Department
University of Petra, Amman, Jordan
mohammed.fasha@uop.edu.jo

Bassam Hammo
King Abdullah II School of Information Technology
University of Jordan, Amman, Jordan
b.hammo@ju.edu.jo

Bilal Sowan
Business Intelligence and Data Analytics Department
University of Petra, Amman, Jordan
bilal.sowan@uop.edu.jo

Husam Barham
Business Intelligence and Data Analytics Department
University of Petra, Amman, Jordan
hbarham@uop.edu.jo

Esam Nsour
Computer Science Department
University of Zarqa, Zarqa, Jordan
ealnsour@zu.edu.jo

**Abstract-** This study uses Jordanian law as a case study to explore the fine-tuning of the Llama-3.1 large language model for Arabic question-answering. Two versions of the model—Llama-3.1-8B-bnb-4bit and Llama-3.1-8B-Instruct-bnb-4bit—were fine-tuned using parameter-efficient fine-tuning (PEFT) with LoRA adapters and 4-bit quantized models, leveraging the Unsloth framework for accelerated and resource-efficient training. A custom dataset of 6000 legal question-answer pairs was curated from Jordanian laws and formatted into structured prompts. Performance was evaluated using the BLEU and the ROUGE metrics to compare the fine-tuned models to their respective base versions. Results demonstrated improved legal reasoning and accuracy while achieving resource efficiency through quantization and optimized fine-tuning strategies. This work underscores the potential of adapting large language models for Arabic legal domains and highlights effective techniques for fine-tuning domain-specific tasks.

## I. Introduction

The increasing complexity of domain-specific queries, such as those related to Arabic legal texts, presents significant challenges for existing large language models (LLMs) [1]. However, developing high-performing Arabic NLP systems remains challenging due to the complexities of the language and the limited availability of annotated datasets, especially in domain-specific areas such as law [2]. The legal domain, in particular, requires models that comprehend natural language and demonstrate reasoning capabilities to answer domain-specific questions accurately.

Large language models like Llama-3.1 have emerged as state-of-the-art tools capable of adapting to various tasks. Fine-tuning these models for specialized applications has proven effective, especially with advancements such as parameter-efficient fine-tuning (PEFT) and quantized model training. These techniques enable efficient adaptation while minimizing computational and memory overhead. Despite these advancements, few studies have explored leveraging large language models for Arabic legal question-answering, mainly using country-specific laws as a case study.

This paper focuses on fine-tuning two variants of Llama-3.1—Llama-3.1-8B-bnb-4bit and Llama-3.1-8B-Instruct-bnb-4bit—for Arabic question-answering tasks within Jordanian law. The models were trained using a custom dataset of legal question-answer pairs, leveraging Unsloth for efficient training and LoRA adapters for parameter-efficient updates. The primary objective is to evaluate and compare the performance of the fine-tuned models with their respective base models to determine their effectiveness in understanding and answering legal questions using external contexts similar to Retrieval-Augmented Generation (RAG) [3].

By addressing these objectives, this study contributes to Arabic NLP by introducing efficient methods for fine-tuning large language models in low-resource domains and providing insights into their applicability in specialized tasks. The results demonstrate the potential of fine-tuned models in delivering accurate and contextually relevant answers, paving the way for practical implementations in legal systems and other specialized fields. This work builds on prior advancements in Arabic NLP, domain adaptation [4]

## II. Related work

### A. Legal NLP

Applying natural language processing (NLP) to legal texts has gained significant attention due to legal language's structured and complex nature. Early works in legal NLP focused on tasks such as information extraction, legal document classification, and case summarization [5]. For Arabic, however, progress has been slower due to the scarcity of high-quality annotated datasets, the unique linguistic challenges posed by its complex grammar, and the coexistence of multiple dialects alongside Modern Standard Arabic [2]. Recent efforts have introduced models like AraBERT and its variations, which are pre-trained for general

Arabic NLP tasks. Still, the model's general architecture makes it suitable for domain-specific tasks when fine-tuned on a relevant dataset [1]. These advancements have paved the way for exploring more sophisticated approaches in Arabic legal domains, such as retrieval-augmented generation.

*B. RAG Techniques*

Retrieval-augmented generation (RAG) is a recent approach that combines document retrieval with generative models to improve the accuracy and relevance of responses for knowledge-intensive tasks [3]. Unlike traditional large language models (LLMs), RAG retrieves relevant documents from a corpus to condition the model's output, proving effective in domains like scientific QA, healthcare, and legal analysis. However, its application in Arabic NLP, especially for legal queries, is limited. This study addresses this gap by fine-tuning Llama-3.1 models for Arabic legal question-answering, providing a foundation for integration into RAG architectures to handle Jordanian legal queries effectively.

The Retrieval-Augmented Generation (RAG) workflow, as illustrated in Figure *1*, involves preparing data by processing and splitting it into chunks, which are converted into embeddings and stored in a vector database. User queries are also transformed into embeddings to search the database for relevant information. The retrieved data is then combined with the query and processed by a large language model (LLM) to produce accurate and contextually relevant responses. The work in this study can be considered an essential step for building offline Arabic legal RAG applications as it aims to prepare offline LLMs for this purpose.

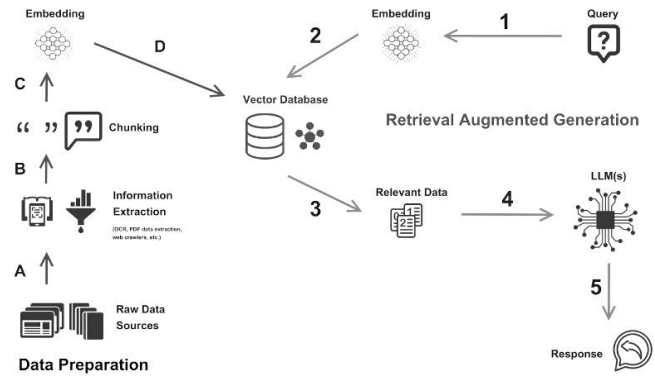

Figure 1 Basic RAG process for information retrieval and response generation. Source: GradientFlow.com

*C. Fine-Tuning for Domain Adaptation*

Fine-tuning large language models on domain-specific datasets has proven a highly effective strategy for improving their performance in specialized applications [7]. For legal texts, fine-tuning allows models to understand complex terminology, syntactic structures, and the distinct language of legal discourse. Studies have shown that fine-tuned LLMs outperform their non-fine-tuned counterparts in tasks like contract analysis, legal judgment prediction, and statutory interpretation [8]. In Arabic, domain adaptation through fine-tuning has been particularly beneficial due to the lack of extensive pretraining data for legal or technical fields [4].

Combining fine-tuning and retrieval-augmented generation can enhance the accuracy and relevance of Arabic legal question answering. This study focuses on the fine-tuning part of the solution.

III. METHODOLOGY

The implemented methodology is shown in Figure 2 below. The work is initiated by collecting and curating Jordanian law documents and extracting relevant articles (مادة قانونية). Then, the extracted legal text was used to create a custom Questions and Answers (QA) dataset by utilizing OpenAI API to generate the (Question, Answer, Context) triples. Later, the PEFT technique was employed to fine-tune a base Llama 3.1 and instruct Llama 3.1 models on the custom dataset. Finally, the fine-tuned models were evaluated using qualitative and quantitative methods to determine the model's suitability for RAG applications.

This research's source code and datasets are publicly available.[1]. Readers can access and use the resources for replication or further exploration. The following presents the implemented methodology and its processing steps.

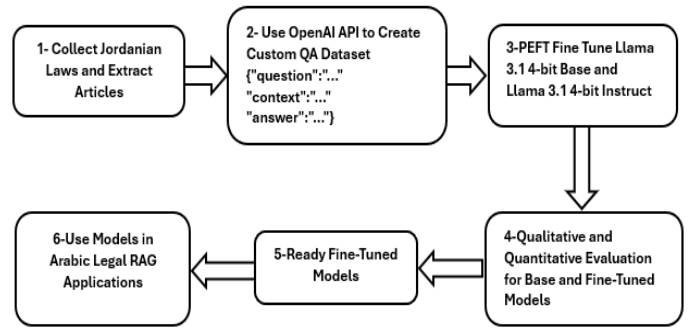

Figure 2-Implementation Methodology

*A. Collecting and Curating the Legal Documents*

The first step involved collecting and curating legal documents from 18 Jordanian laws, namely the laws shown in Table below.

Table 1 The 18 Jordanian Laws Used to Fine-Tune the Models

| # | Selected Jordanian Laws |
|---|---|
| 1 | Jordanian Civil Law<br>القانون المدني الأردني |
| 2 | Electronic Crimes Law 2024<br>قانون الجرائم الإلكترونية 2024 |
| 3 | Jordanian Labor Law with Full Amendments<br>قانون العمل الأردني مع كامل التعديلات |
| 4 | Narcotics and Psychotropic Substances Law<br>قانون المخدرات والمؤثرات العقلية |
| 5 | Principles of Criminal Court Law<br>قانون أصول المحاكمات الجزائية |
| 6 | Principles of Civil Court Procedures Law<br>قانون أصول المحاكمات المدنية |
| 7 | Jordanian Data Law<br>قانون البيانات الأردني |
| 8 | Arbitration Law<br>قانون التحكيم الأردني |

---

[1] https://github.com/msfasha/Research-Resources/tree/main/ArabicLegalLLM

| 9 | Jordanian Execution Law |
| | قانون التنفيذ الأردني |
| 10 | Jordanian Economic Crimes Law |
| | قانون الجرائم الاقتصادية الأردني |
| 11 | Jordanian Companies Law |
| | قانون الشركات الأردني |
| 12 | Jordanian Penal Code |
| | قانون العقوبات الأردني |
| 13 | Formation of Regular Courts Law |
| | قانون تشكيل المحاكم النظامية |
| 14 | Law on the Protection of Copyright |
| | قانون حماية حق المؤلف |
| 15 | Conciliation Courts Law |
| | قانون محاكم الصلح |
| 16 | Law on the Trial of Ministers |
| | قانون محاكمة الوزراء |
| 17 | Bar Association Law, including regulations for agents |
| | قانون ونظام وتعليمات نقابة المحامين والوكلاء |
| 18 | Law, Regulation, and Instructions for Commercial Agents and Brokers |
| | قانون ونظام وتعليمات الوكلاء والوسطاء التجاريين |

The above laws were collected and curated, and their articles were extracted to prepare for the next step, where the (questions, answers, context) triples are created. 3578 articles were extracted from these 18 laws.

B. Creating (Question, Answer, Context) Triples

OpenAI API with GPT-3.5 Turbo was utilized to extract question-answer and context triples from the curated legal articles. The generated dataset comprised around 6,000 QA pairs, and the context was used to create a question and answer based on each legal article. These QA pairs and context were reviewed to ensure contextual and semantic accuracy, forming a robust dataset for Arabic legal question-answering.

The dataset was split into training (80%) and testing (20%) subsets to facilitate model fine-tuning and evaluation. The training subset was used to adapt the models to the legal domain, while the testing subset was reserved for evaluating their performance on unseen data.

C. Model Fine-Tuning

Two pre-trained Llama-3.1 models—Llama-3.1-8B-bnb-4bit and Llama-3.1-8B-Instruct-bnb-4bit—were selected for fine-tuning. Both models utilize 4-bit quantization, which reduces memory usage and speeds up training, making them suitable for resource-efficient adaptation to domain-specific tasks.

The fine-tuning process was conducted in a Google Colab Pro environment with an NVIDIA A100-SXM4-40GB GPU. This GPU's high memory capacity (39.564 GB total, with 6.004 GB reserved during training) and computational power allowed for efficient training of the models using large datasets, even with the computational overhead introduced by LoRA adapters and PEFT.

Fine-tuning steps included:

- Adapting the models using LoRA adapters with a configuration of r=16, lora_alpha=16, and no dropout.

- Optimizing the models with a learning rate of 2e-4, a batch size of 2 (using gradient accumulation for effective larger batch sizes), and FP16 or BF16 precision, depending on hardware support.

- Training for one epoch, balancing computational efficiency with practical learning.

The Colab environment's support for modern GPUs and ease of access made it a practical choice for this fine-tuning process, particularly for leveraging 4-bit quantized models without running into memory limitations.

IV. RESULTS AND ANALYSIS

A. Quantitative Analysis of Results

The quantitative evaluation compares the performance of the four non-fine-tuned and fine-tuned generator models. As shown in Table *1* below, the fine-tuned models outperformed the base models across both BLEU and ROUGE metrics.

Table 1- Performance Results

| Model | BLEU | ROUGE-1 | ROUGE-2 | ROUGE-L |
|---|---|---|---|---|
| llama3.1 | 0.058 | 0.026 | 0.000 | 0.026 |
| llama3.1_instruct | 0.128 | 0.070 | 0.013 | 0.070 |
| fine_tuned_llama3.1 | 0.290 | 0.081 | 0.020 | 0.081 |
| fine_tuned_llama3.1_instruct | 0.270 | 0.063 | 0.005 | *0.063* |

Both fine-tuned models (fine_tuned_llama3.1 and fine_tuned_llama3.1_instruct) achieved higher BLEU and ROUGE scores compared to their respective base models (llama3.1 and llama3.1_instruct). This indicates that the fine-tuning process has improved the models' ability to generate more similar responses to the human-written ground truth.

**Impact of Instruction Tuning:**

The fine-tuned models improve over the base models, suggesting that instruction tuning can effectively enhance the model's performance on this dataset. However, when comparing the fine-tuned versions, fine_tuned_llama3.1 slightly outperforms fine_tuned_llama3.1_instruct regarding BLEU and ROUGE scores.

**Possible Interpretations:**

- **Fine-tuning Effectiveness:** The significant improvement in BLEU and ROUGE scores for the fine-tuned models highlights the effectiveness of fine-tuning on this specific dataset and task. Fine-tuning allows the model to learn task-specific patterns and improve its ability to generate relevant and accurate responses.

- **Comparing the Performance of Fine-Tuned Models:** The results show that the fine-tuned Llama 3.1 model slightly outperformed the fine-tuned Llama 3.1 Instruct, despite the latter being pre-trained for instruction-following tasks. This might occur because a general-purpose model like Llama 3.1 starts with a more neutral initialization, allowing fine-tuning to shape the model specifically to the target task, such as Arabic legal question-answering. In contrast, Llama 3.1 Instruct might have prior biases introduced by its instruction-tuning process, which could conflict with the patterns in the fine-tuning dataset. In addition, the slight differences in BLEU and ROUGE scores could also stem from overfitting or differences in training dynamics, where the general-purpose model achieved a more balanced adaptation. These findings suggest that the effectiveness of fine-tuning depends not only on the base model's capabilities but also on the alignment between the dataset and the model's pre-training objectives.

**Further Analysis:**

- **Human Evaluation:** While BLEU and ROUGE scores provide quantitative measures, it's crucial to complement them with human evaluation to assess the quality of the generated responses regarding fluency, coherence, and relevance.

- **Error Analysis:** Analyzing the specific cases where the models generate responses with low BLEU and ROUGE scores can provide insights into the types of errors they make and potential areas for improvement.

- **Dataset Exploration:** Examining the dataset's characteristics, such as the types of questions, the complexity of the responses, and the presence of instructions, can help explain the observed differences in model performance. Also, the dataset can be reformatted to test the impact of clearer instruction-based inputs and further hyperparameter tuning for both models to maximize their performance.

In summary, the results indicate that fine-tuning effectively improves the performance of the language models on this dataset, and instruction tuning can provide additional benefits. However, further analysis and human evaluation are necessary to gain a deeper understanding of the strengths and weaknesses of each model.

*B. Qualitative Analysis of Results*

This section complements the quantitative analysis by qualitatively evaluating the performance of the different models on the dataset, focusing on their ability to generate accurate and contextually relevant responses. For this purpose, we randomly selected four samples at the head of the random test-split dataset.

**General Trends in Model Performance**

The fine-tuned models, fine_tuned_llama3.1 and fine_tuned_llama3.1_instruct, consistently demonstrated superior performance to the base models (llama3.1 and llama3.1_instruct). Fine-tuning significantly enhanced their ability to comprehend the contextual requirements of legal and procedural queries, resulting in more accurate and concise responses. By contrast, the base models often generated verbose and generalized answers less aligned with the ground truth.

For example, in case-1, which is related to procedural timing of requests ("متى يُقبل طلب الرد؟"), the ground truth specifies that the request should be submitted before entering the trial or proceedings. Fine-tuned models provided answers that were closely aligned with this procedural requirement. Fine_tuned_llama3.1 concisely stated the condition, while fine_tuned_llama3.1_instruct included a plausible alternative procedural nuance. On the other hand, base models failed to adequately capture this specificity, with responses either diverging into verbose explanations or misrepresenting procedural details. Similarly, in case 2, the fine-tuned models identified key urgent measures, such as detention and release, reflecting the ground truth. In contrast, base models, particularly llama3.1_instruct, struggled with this question, instead providing vague generalities unrelated to the procedural context. In the same respect, the fine-tuned models correctly focused on the court's responsibility to evaluate evidence and weigh testimony, as outlined in the ground truth. These responses demonstrated contextual precision and brevity, while the base models either generalized the court's role or introduced extraneous details, diluting their answers' relevance. Finally, in case 4, the fine-tuned models excelled in identifying both critical conditions for pardon: reporting the crime before its discovery and returning the stolen funds, whereas base models tended to emphasize only one condition or include unrelated procedural details, showcasing a lack of contextual depth.

The fine-tuned models generally demonstrated notable improvements in handling domain-specific tasks, particularly in legal and procedural contexts. They consistently provided more accurate and contextually relevant responses than the base models, underscoring the efficacy of fine-tuning in enhancing language model performance. However, opportunities remain to refine the models, particularly in handling complex queries and balancing detail with precision. These findings align with quantitative analysis and highlight the potential for continued optimization in specialized domains.

Table 2-Randomly Selected Samples for Qualitative Analysis

| ID | Question | Ground Truth | Llama 3.1 | Llama 3.1 Instruct | Fine-tuned Llama 3.1 | Fine-tuned Llama 3.1 Instruct |
|---|---|---|---|---|---|---|
| 2969 | متى يُقبل طلب الرد؟ | يُقبل طلب الرد إذا تم تقديمه قبل الدخول في الدعوى أو المحاكمة. | The question is asking when the request for a reconsideration is accepted. The input provides context for the question, which is a law in Jordan. | يُقبل طلب الرد بتقديم استدعاء إلى رئيس محكمة البداية إذا كان المطلوب رده قاضي صلح أو أحد قضاة المحكمة البدائية أو إلى رئيس محكمة الاستئناف إذا كان قاضي استئناف أو رئيس. | يُقبل طلب الرد قبل الدخول في الدعوى أو المحاكمة. | يُقبل طلب الرد في أول جلسة تلي الحادث إذا كان سبب الرد متولداً عن حدث طرأ بعد الدخول في الدعوى أو المحاكمة. |
| 2284 | ما هو الإجراء الذي يجوز اتخاذه حتى مع وقف الدعوى وفقاً للمادة 140؟ | يجوز اتخاذ الإجراءات والتدابير الضرورية أو المستعجلة مثل التوقيف وإخلاء السبيل حتى مع وقف الدعوى. | ما هو الإجراء الذي يجوز اتخاذه حتى مع وقف الدعوى وفقاً للمادة 140؟ الإجراء الذي يجوز اتخاذه حتى مع وقف الدعوى وفقاً للمادة 140 هو إجراء توقيف وإخلاء السبيل. | الإجابة الصحيحة هي: أن توقف الدعوى. لأن المادة 140 من قانون الاختصاصات تنص على أن يجوز للمحكمة وقف الدعوى حتى يتم الفصل في دعوى أخرى. وتحدد المادة 140 في فقرتها. | يجوز للمحكمة أن توقف الدعوى وتحدد للمشتكى عليه أو للمدعي بالحق الشخصي أو للمجني عليه أجلًا لرفع المسألة إلى المحكمة المختصة. | اتخاذ الإجراءات الضرورية أو المستعجلة كالتوقيف وإخلاء السبيل. |
| 3177 | ما هو دور المحكمة وفقًا للمادة 34؟ | دور المحكمة وفقًا للمادة 34 هو ترجيح البينة بين الأدلة والشهادات في الدعوى. | المحكمة هي الهيئة التي تقوم بمهمة التحكيم ووضع الأحكام. | ينبغي للمحكمة أن ترجح بينة على أخرى وفقاً لما تستخلصه من ظروف الدعوى، كما لا يجوز لها أن تصدر حكما في أية قضية بالاستناد الى شهادة شاهد فرد إلا إذا لم يعترض عليها ال. | تستطيع المحكمة أن ترجح بينة على أخرى وفقًا للمادة 34. | المحكمة تُستخدم في حالات الرجحان بين بينة على أخرى وفقًا للمادة 34. |
| 3536 | ما هو شرط العفو من العقوبة المقررة للجرائم التي يكون محلها مالا عاما؟ | التبليغ عن الجريمة قبل اكتشافها ورد المال محل الجريمة. | يجب على المحكمة أن تقضي بالحد الأدنى للعقوبة، إذا كانت الجريمة التي ارتكبتها الشركاء قد يكون محلها مالا عاما، وتم اكتشافها قبل صدور الحكم النهائي بها، وتم التبليغ عنها. | يجب أن يكون التبليغ قبل اكتشاف الجريمة. إذا حصل التبليغ بعد اكتشاف الجريمة وقبل صدور الحكم النهائي بها، فيجب على المحكمة أن تقضي بالحد الأدنى للعقوبة ويجب عليها أخ. | يجب أن يكون التبليغ قبل اكتشاف الجريمة. | يجب أن يكون التبليغ قبل اكتشاف الجريمة وينتج عنه رد المال محل الجريمة. |

## C. Future Work

To build on the findings of this study, several avenues for future research are proposed:

- **Testing Alternative Retrieval Models**:

  This work represents a step toward applying fine-tuned models in Retrieval-Augmented Generation (RAG) pipelines. The next critical step is to test these fine-tuned models within a complete RAG environment where the input context is not curated and may include unstructured or semi-structured legal content, such as court rulings, legal memos, or annotated contracts. Exploring how the fine-tuned models handle such diverse and unstructured inputs will provide insights into their robustness and adaptability in real-world applications.

- **Dataset Expansion and Model Diversification**

  The current dataset primarily focuses on Modern Standard Arabic and Jordanian legal texts, providing a strong foundation for fine-tuning language models in a specific legal context. Future efforts should prioritize expanding this dataset to include a broader range of Arabic dialects, legal systems, and unstructured content such as court transcripts, legal memos, and case law. This expansion will improve the model's generalization and adaptability across diverse Arabic-speaking regions and legal scenarios.

  In addition, testing other state-of-the-art language models, including different architectures and scales, could provide valuable insights into their comparative performance on Arabic legal question-answering tasks. Exploring models trained on larger corpora, multilingual models, or those explicitly designed for low-resource languages can help identify the most suitable candidates for this domain.

  By addressing these limitations and pursuing the proposed future directions, this research can contribute to developing more robust, accurate, and versatile systems for answering Arabic legal questions.

## V. CONCLUSION

This study demonstrates the effectiveness of fine-tuning large language models for Arabic legal question-answering, specifically within Jordanian law. The models produced contextually accurate and legally relevant answers by curating a domain-specific dataset and employing parameter-efficient fine-tuning techniques, outperforming their non-fine-tuned counterparts. The findings underscore the critical role of fine-tuning in adapting large language models to specialized domains, addressing challenges like specialized terminology and linguistic structures.

This research contributes to the advancement of Arabic NLP by offering a replicable methodology for applying LLMs to legal domains. It contributes to setting foundations for further innovation in Arabic legal question-answering systems, paving the way for practical applications such as legal research assistants, automated document analysis, and enhanced legal education tools.